\documentclass{article}

    \usepackage[preprint]{neurips_2024}

\usepackage[utf8]{inputenc} % allow utf-8 input
\usepackage[T1]{fontenc}    % use 8-bit T1 fonts
\usepackage{hyperref}       % hyperlinks
\usepackage{url}            % simple URL typesetting
\usepackage{booktabs}       % professional-quality tables
\usepackage{amsfonts}       % blackboard math symbols
\usepackage{nicefrac}       % compact symbols for 1/2, etc.
\usepackage{microtype}      % microtypography
\usepackage{xcolor}         % colors
\usepackage{graphicx} % for images
\usepackage{subcaption} % this is for graphs 
\usepackage{float} % This is to force images

\title{Fine-Tuning Language Models for Ethical Ambiguity: A Comparative Study of Alignment with Human Responses}

\author{%
Pranav Senthilkumar \quad Visshwa Balasubramanian \quad Prisha Jain \\ \textbf{Aneesa Maity} \quad \textbf{Jonathan Lu} \quad \textbf{Kevin Zhu}
\\
Algoverse AI Research\\
\texttt{jonathan@algoverse.us, kevin@algoverse.us}
}

\begin{document}

\maketitle

\begin{abstract}
  
Language models often misinterpret human intentions due to their handling of ambiguity, a limitation well-recognized in NLP research. While morally clear scenarios are more discernible to LLMs, greater difficulty is encountered in morally ambiguous contexts. In this investigation, we explored LLM calibration to show that human and LLM judgments are poorly aligned in such scenarios. We used two curated datasets from the Scruples project for evaluation: \textbf{DILEMMAS}, which involves pairs of distinct moral scenarios to assess the model’s ability to compare and contrast ethical situations, and \textbf{ANECDOTES}, which presents individual narratives to evaluate the model’s skill in drawing out details, interpreting, and analyzing distinct moral scenarios.
Model answer probabilities were extracted for all possible choices and compared with human annotations to benchmark the alignment of three models— Llama-3.1-8b, Zephyr-7b-beta, and Mistral-7b.
Significant improvements were observed after fine-tuning, with notable enhancements in both cross-entropy and Dirichlet scores, particularly in the latter. Notably, after fine-tuning, the performance of Mistral-7B-Instruct-v0.3 was on par with GPT-4o. However, the experimental models that were examined were all still outperformed by the BERT and RoBERTa models in terms of cross-entropy scores\citep{lourie2021scruplescorpuscommunityethical}.  Our fine-tuning approach, which improves the model’s understanding of text distributions in a text-to-text format, effectively enhances performance and alignment in complex decision-making contexts, underscoring the need for further research to refine ethical reasoning techniques and capture human judgment nuances.
\end{abstract}

\section{Introduction}
Language models, despite their strong capabilities in generating human-like text, still face inconsistent alignment with human decision-making in ambiguous scenarios. While Reinforcement Learning from Human Feedback (RLHF) guides models toward human-preferred outcomes, it does not fully address the inherent subjectivity in morally complex situations due to the variability in human values, the complexity of moral reasoning, and the limitations in feedback and representation \citep{RLHF-Citation}. This leaves room for inconsistency in model outputs, especially when human judgments are nuanced and subjective. Ambiguity in decision-making occurs when outcomes are equally favorable or unfavorable, making individuals rely on heuristics and bias to make final decisions\citep{Heuristics}.LLMs leverage large datasets to learn patterns and handle tasks involving ambiguous inputs. \citet{brown2020languagemodelsfewshotlearners} highlight that while these models generate relevant text, they often rely on surface-level patterns rather than comprehending deeper semantics. This challenge is evident in tasks like open-domain question answering, where ambiguity in entity references or events can lead to multiple plausible interpretations. Although LLMs perform well in generating coherent responses, their ability to disambiguate contextual queries, especially in morally ambiguous scenarios, remains under explored \citep{scherrer2023evaluatingmoralbeliefsencoded}. 
The goal of this investigation is to determine whether language models can effectively replicate human collective moral judgments or if they exhibit inherent biases. To explore this, we analyze the next token probability distributions to understand how well these models align with or diverge from human decision-making in ambiguous moral contexts. Understanding the similarities and differences can help improve LLM design, making them more reliable in real-world applications where ambiguity is often present. 

Our contributions are as follows:
\begin{itemize}
    \item \textbf{Investigation and Evaluation of Ambiguity:}  We demonstrate that LLMs are not representative of diverse moral preferences when presented with complex and nuanced moral scenarios. Furthermore, we find that fine-tuning on response distributions in the text is effective and improves alignment with moral perspectives. 
\end{itemize}

\section{Related Works}
\subsection{Moral Values In LLMs}
A significant body of literature has sought to explore the presence of moral values in LLMs and how those values align with human beliefs. This has been accomplished by assessing LLMs through a variety of moral perspectives. For instance, past literature has evaluated LLMs as survey respondents in both low-and high-ambiguity scenarios, based on the GERT morality framework, which outlines ten rules that form the basis of common sense morality \citep{scherrer2023evaluatingmoralbeliefsencoded}. This study found that in clear scenarios model chose the moral answer and vice versa.

Other studies have also evaluated LLMs’ ability to predict human behavior, using five basic ethics perspectives: justice, virtue, deontology, utilitarianism, and commonsense \citep{hendrycks2023aligningaisharedhuman}. Here, rather than ambiguous scenarios, clear-cut situations with definite answers were used to evaluate the models, along with human annotations for training. This study also exposed the LLMs weakness in this domain. Although the alignment of LLMs’ expressed morals with human values has been studied, LLM calibration in morally ambiguous scenarios has not.
\subsection{LLM Calibration in Factual Contexts}
Calibration is a measure of the trustworthiness of an LLM, comparing the confidence scores output by the LLM to the ground truth values. It allows users to see whether the LLM has an accurate gauge of its uncertainty \citep{kassner2023languagemodelsrationality}.

This study is similar to the aforementioned studies in that it also evaluates the morality of LLMs. However, our contribution involves evaluating the models’ calibration with respect to human annotators’ answers by modeling output distributions. This allows for a more comprehensive insight into whether a model is truly aligned with the user population.

Existing research on calibration has been fact-based. One study evaluated the effect of various changes made during the training and construction phases on the calibration of LLMs for causal language modeling, fact generation, and multi-language understanding. It was revealed that larger parameter scales and longer training dynamics during pre-training improve calibration, while instruction tuning and synthetic data deteriorate it.  \citep{kassner2023languagemodelsrationality}. 
\subsection{LLM Calibration In Subjective Contexts}
Although most research on LLM calibration has been fact-based, there have been some studies that investigate LLM calibration in subjective contexts
An alignment study, for instance, used calibration to assess the alignment of LLM responses with the opinions of various demographic groups in the US. It tested the models on various political topics and also evaluated whether models could better represent certain demographics after being steered. It found that models fine-tuned with human feedback are generally left-leaning and that steering models to represent certain underrepresented demographics did not significantly improve their abilities to answer as those demographics \citep{scherrer2023evaluatingmoralbeliefsencoded}.

Though we applied a similar methodology to gauge alignment, our focus is exclusively on ambiguous moral situations. Rather than evaluating whether an LLM can accurately represent the political opinions of a diverse population, we assessed whether LLMs can represent the moral values and judgments of a population in morally contentious situations.

\section{Methodology}
\maketitle

\subsection{Model Calibration Approach}
To measure the calibration of LLM responses with respect to human ethical judgments, we extracted token probabilities from each LLM's final softmax layer.

\subsection{Datasets}

\textbf{}Two primary datasets were used to facilitate this measurement: the Anecdotes dataset and the Dilemmas dataset, both derived from \citep{lourie2021scruplescorpuscommunityethical}. These datasets provide ethical judgments based on real-world scenarios, allowing us to compare LLM predictions against collective human judgments.

\textbf{Anecdotes:} The Anecdotes dataset comprises 32,000 real-life scenarios where individuals seek ethical judgments from the online subreddit 'r/AmItheAsshole'. For simplicity, we adapted this dataset by converting the ethical judgments into a binary decision task. Specifically, each judgment is categorized into classes such as "AUTHOR," "OTHER," "EVERYBODY," "NOBODY," and "INFO," representing who the community believes is in the wrong. To derive the \textbf{Binarized Label Scores}, responses are normalized to estimate the probability of each view by counting votes for each class, converting these counts into probability distributions, and then binarizing the labels into "RIGHT" or "WRONG" categories based on majority consensus. Furthermore, to pass in the bulk of the text, we provide the \textbf{title}, \textbf{text}, and \textbf{action} taken by the individual to the model, along with a few shot prompt to guide the model to only output 'YES' or 'NO'. The model's predictions are then compared against these binarized labels to evaluate its performance.

\textbf{Dilemmas:} The Dilemmas dataset contains 10,000 ethical dilemmas, which were annotated by crowd sourcing through Mechanical Turk. The dilemmas themselves were collected separately, and the crowd sourcing process focused on annotating these scenarios in terms of paired actions. The task was to identify which of the two actions was less ethical. For this dataset, we filtered the two actions and appended a few-shot prompt in order to assist the model in its probability generation. We used the 'gold-annotations' provided in the dataset as the ground truth or human probabilities

\subsection{Model Selection}
We evaluated four different LLMs: \textbf{GPT4o}, \textbf{Llama-3.1-8B}, \textbf{Zephyr-7B-Beta}, and \textbf{Mistral-7B}.GPT4o was chosen as a baseline due to its established performance in ethical judgment tasks\citep{GPT4o}.

\subsection{Loss Functions for Calibration Measurement}
To measure the alignment between the model's predictions and human judgments, we employed Binary Cross-Entropy Loss and Dirichlet Multinomial Loss:
\begin{itemize}
    \item \textbf{Binary Cross-Entropy Loss}: This quantifies the discrepancy between the predicted probability distribution and binary labels. In the context of soft cross-entropy, as discussed by Scruples, the loss is computed using an empirical Bayesian approach (Murphy, 2012). The prior $\alpha$ is estimated via maximum likelihood, denoted as $\hat{\alpha}$, and the expected loss is determined over the posterior distribution. Specifically, for soft labels, the loss is calculated as:

\[
s = \mathbb{E}_{p|Y, \hat{\alpha}} \left[ \sum_{i} \sum_{j} Y_{ij} \log p_{ij} \right]
\]

where $p_{ij}$ represents the predicted probability for the $j$-th class and $Y_{ij}$ denotes the corresponding soft label for the $i$-th instance. \citep{lourie2021scruplescorpuscommunityethical}
    \item \textbf{Dirichlet-Multinomial Loss}: This loss function extends binary cross-entropy by incorporating a Dirichlet prior to measuring the discrepancy between predicted probabilities and actual outcomes. It provides a refined evaluation by modeling class distributions rather than single probability estimates.

\end{itemize}

\section{Experiments}

\begin{figure}[H]
    \centering
    \includegraphics[width=\linewidth]{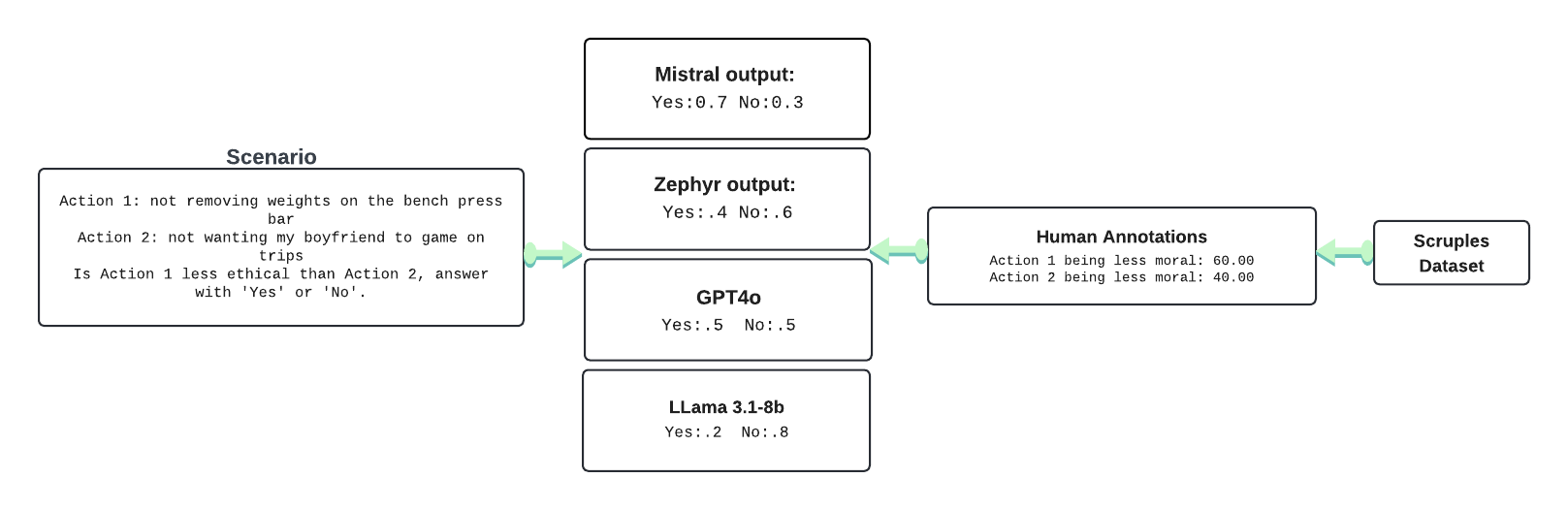} % Adjust width to fit within page limits
    \caption{Evaluation process outline for the Dilemmas Dataset.}
    \label{fig:experiment-image}
\end{figure}

\subsection{Training Process and QLoRA Fine-Tuning}
Fine-tuning employed the QLoRA \citep{dettmers2023qloraefficientfinetuningquantized} technique, utilizing the training splits from both datasets. This method aligned model predictions with human ethical judgments while maintaining memory efficiency.Post fine-tuning, the models were evaluated on the test splits of both datasets. We compared next-token probabilities to human judgments using Cross-Entropy and Dirichlet Multinomial Loss to measure congruence between model predictions and human decisions. The results were quantified by averaging these losses for each model across the Anecdotes and Dilemmas datasets.

\section{Results}

% First Table: Dilemmas Model Performance Comparison
\begin{table}[H] % Changed positioning to H to force placement exactly here
\centering
\caption{Dilemmas Model Performance Comparison}
\renewcommand{\arraystretch}{1.2} % Increase row height
\begin{tabular}{lcc|cc}
\toprule
\textbf{Model Name} & \multicolumn{2}{c}{\textbf{\vspace{0.2cm} Cross-Entropy Loss}} & \multicolumn{2}{c}{\textbf{\vspace{0.2cm} Dirichlet Loss}} \\
                    & \textbf{Original} & \textbf{Finetuned} & \textbf{Original} & \textbf{Finetuned} \\
\midrule
Zephyr-7b-beta          & 0.7316  & 0.6991 (\textbf{-4.44\%}) & 4.702  & 3.333 (\textbf{-29.12\%}) \\
Mistral-7B-Instruct-v0.3 & 0.7088  & 0.6999 (\textbf{-1.26\%}) & 4.508  & 3.214 (\textbf{-28.70\%}) \\
Meta-Llama-3-8B-Instruct & 0.7431  & 0.6837 (\textbf{-7.99\%}) & 3.452  & 3.287 (\textbf{-4.78\%}) \\
\bottomrule
\end{tabular}
\end{table}

% Second Table: Anecdotes Model Performance Comparison
\begin{table}[H] % Use H to enforce the table placement exactly after the first table
\centering
\caption{Anecdotes Model Performance Comparison}
\renewcommand{\arraystretch}{1.2} % Increase row height
\begin{tabular}{lcc|cc}
\toprule
\textbf{Model Name} & \multicolumn{2}{c}{\textbf{\vspace{0.2cm} Cross-Entropy Loss}} & \multicolumn{2}{c}{\textbf{\vspace{0.2cm} Dirichlet Loss}} \\
                    & \textbf{Original} & \textbf{Finetuned} & \textbf{Original} & \textbf{Finetuned} \\
\midrule
Zephyr-7b-beta          & 0.6971  & 0.6501 (\textbf{-6.74\%}) & 12.9413 & 8.9426 (\textbf{-30.90\%}) \\
Mistral-7B-Instruct-v0.3 & 0.6695  & 0.6654 (\textbf{-0.61\%}) & 12.5013 & 8.5354 (\textbf{-31.72\%}) \\
Meta-Llama-3-8B-Instruct & 0.8527  & 0.6010 (\textbf{-29.52\%}) & 10.8926 & 8.2331 (\textbf{-24.42\%}) \\
\bottomrule
\end{tabular}
\end{table}

\subsection{Non-Finetuned Dilemmas}

 Across all tested models for the Dilemmas dataset, our baseline (GPT-4o) exhibited the lowest cross-entropy of 0.6691, followed by the three experimental models with cross-entropy values around 0.725 ± 0.2.  GPT-4o’s more polarized outputs also support this; GPT-4o displayed a 90\% chance of either outputting “Yes” or “No” regardless of the scenario, while the other models showed more variance in their outputs. Although the variance in cross-entropy averages provides a more nuanced understanding of model performance, it is important to note that all models had a relatively similar range of cross-entropy values.

In contrast, the average Dirichlet multinomial loss values for GPT-4o were the highest compared to the other three models. The Dirichlect loss, focusing on the model's calibration state, indicates that LLama was the most calibrated to human preferences, while GPT was the least calibrated.

Although infrequently, with some scenarios, a total flip in model and human probabilities inclination towards certain answer choices was observed. For example, if the human annotations indicated the second action to be less ethical, the model demonstrated a high accuracy in the first action being less ethical.

In practice, this was observed with the following scenario: 

Action 1: posting amateur work on r/piano
Action 2: being mad at my boyfriend for not trusting me about being where I say I am
Which action is less ethical? Yes for Action 1 or No for Action 2?

\textbf{Mistral Output:}

\small % Reduce the font size
\begin{ttfamily}
Yes Probability (percent): 94.78024266842719 \\
No Probability (percent): 5.2197573315728025
\end{ttfamily}
\normalsize % Reset to normal size

\textbf{Zephyr Output:}

\small % Reduce the font size
\begin{ttfamily}
Yes Probability (percent): 98.15524597187405 \\
No Probability (percent): 1.8447540281259505
\end{ttfamily}
\normalsize % Reset to normal size

\textbf{Llama 3.1-8b Output:}

\small % Reduce the font size
\begin{ttfamily}
Yes Probability (percent): 75.4035923356711 \\
No Probability (percent): 24.596407664328883
\end{ttfamily}
\normalsize % Reset to normal size

\textbf{GPT-4 Output:}

\small % Reduce the font size
\begin{ttfamily}
Yes Probability (percent): 0.19\% \\
No Probability (percent): 99.80\%
\end{ttfamily}
\normalsize % Reset to normal size

\textbf{Human Annotations}

\small % Reduce the font size
\begin{ttfamily}
Human Right Probability (percent): 20.00\% \\
Human Wrong Probability (percent): 80.00\%
\end{ttfamily}
\\

Although GPT-4 excelled in this scenario, there were numerous instances where all models struggled and deviated significantly from human annotations, even occasionally showing a completely opposite distribution. This could be due to how models generalize from large datasets, as "high-capacity models... begin to learn how to perform a surprising amount of tasks without the need for explicit supervision." However, in morally complex scenarios, their reasoning may rely more on dataset biases than on true ethical understanding\citep{Radford2019}

\subsection{Non-Finetuned Anecdotes}:
In anecdotal scenarios, Zephyr-7b-beta and GPT-4o performed comparably well, indicating robustness in handling these cases. Mistral also showed improved performance on this dataset relative to the Dilemmas dataset, suggesting that its fine-tuning might have had a positive effect. Conversely, Llama 3.1-8b demonstrated notably poorer performance, which may indicate limitations in its ability to capture the nuances of the anecdotes effectively. 

Overall, Mistral 7b, Zephyr 7b-beta, and GPT-4o all exhibit strong performance on this dataset, indicating robustness across various types of texts.  On the other hand, Llama’s weaker performance suggests that anecdotal scenarios are more difficult than short dilemmas, possibly due to the variability.

On the other hand, all models display a considerable increase in Dirichlet Multinomial loss values. While in dilemmas, the losses were around 3-5, the losses for the anecdotes were all in the low tens. This further suggests that these models are not calibrated well to deal with the narrative complexity of such anecdotal scenarios compared to the more straightforward nature of dilemmas. One possible explanation is that the crowdsourced Anecdotes often involved subtle context clues that made the situations more open-ended, making it harder for the models to align their probabilities with human annotations.

In summary, for Dilemmas, the models are less confident (higher cross-entropy) but better calibrated (lower Dirichlet), as the scenarios in this dataset are structured such that the models can better align their predictions with real-world outcomes. On the other hand, in the Anecdotes scenarios, the models are more confident (lower cross-entropy) but often miscalibrated (higher Dirichlet), likely due to the complexity and variability in narrative contexts. 

\subsection{Finetuned Dilemmas}
After fine-tuning, the Zephyr-7b-beta model achieved a cross-entropy score of 0.6991 and a Dirichlet loss of 3.333, both improved from its initial values. The Mistral-7B-Instruct-v0.3 model also showed better performance with a cross-entropy score of 0.6699 and a Dirichlet loss of 3.214. These improvements indicate that fine-tuning enhanced the models' ability to better match the true probability distributions of ethical judgments. However, the reductions were modest, suggesting potential overfitting or a need for further optimization in the fine-tuning process.
\subsection{Finetuned Anecdotes}
For the Anecdotes dataset, the fine-tuned models demonstrated mixed results. The Llama-3.1-8B model achieved a cross-entropy score of 0.6837, and Zephyr-7b-beta had a score of 0.6991. While cross-entropy scores remained relatively stable, Dirichlet losses improved significantly, with Llama-3.1-8B at 3.287 and Zephyr-7b-beta at 3.333. This suggests that fine-tuning enhanced model calibration for handling narrative complexity, though the Dirichlet losses remain higher compared to the Dilemmas dataset, reflecting the greater challenge of the anecdotal data.

\begin{figure}[H]
    \centering
    \begin{subfigure}[t]{0.43\textwidth} % Adjusted width to fill more space
        \centering
        \includegraphics[width=\linewidth]{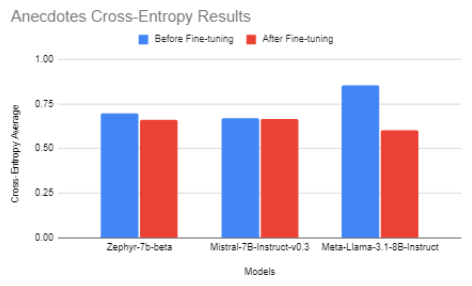}
        \caption{Anecdotes Cross Entropy}
        \label{fig:anecx}
    \end{subfigure}
    \hspace{0pt} % Removes space between subfigures
    \begin{subfigure}[t]{0.43\textwidth} % Adjusted width to fill more space
        \centering
        \includegraphics[width=\linewidth]{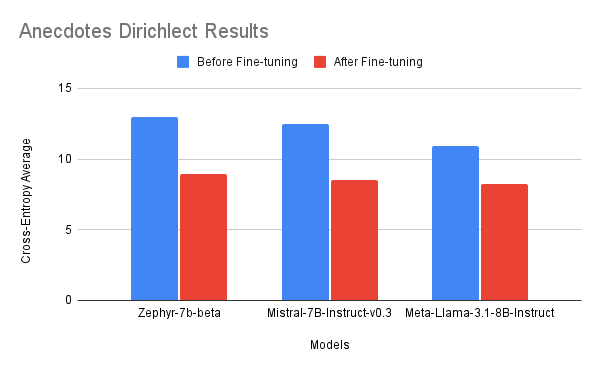}
        \caption{Anecdotes Dirichlet}
        \label{fig:anecd}
    \end{subfigure}
    \caption{Metrics for Anecdotes dataset showing alignment between models and human judgment}
    \label{fig:anecdotes}
\end{figure}

\begin{figure}[H]
    \centering
    \begin{subfigure}[t]{0.4\textwidth}
        \centering
        \includegraphics[width=\linewidth]{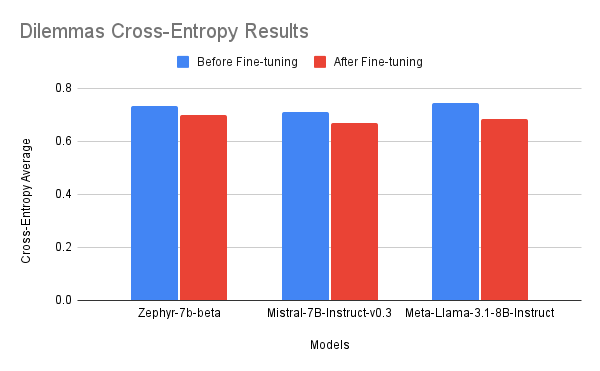}
        \caption{Dilemmas Cross Entropy}
        \label{fig:dilx}
    \end{subfigure}
    \hfill
    \begin{subfigure}[t]{0.4\textwidth}
        \centering
        \includegraphics[width=\linewidth]{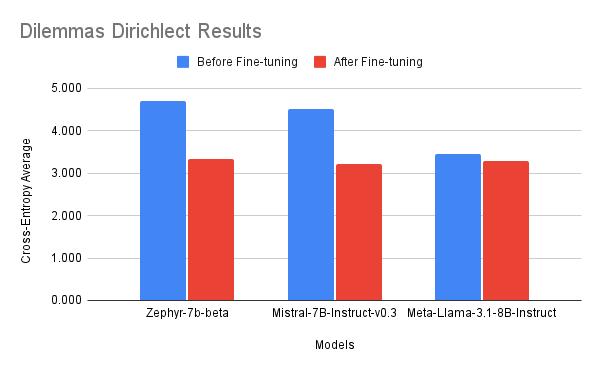}
        \caption{Dilemmas Dirichlet}
        \label{fig:dild}
    \end{subfigure}
    \caption{Metrics for Dilemmas dataset showing alignment between models and human judgment}
    \label{fig:dilemmas}
\end{figure}

\section{Conclusion}

In summary, fine-tuning led to different outcomes based on the datasets: notable progress on the Dilemmas dataset but stronger performance on Anecdotes, where models showed increased confidence in accurately reflecting human opinions in more open-ended narrative tasks. This study underscores how the nature of the dataset influences the effectiveness of fine-tuning, revealing that while our approach significantly improved model performance and alignment, persistent calibration issues remain. The findings highlight a critical need for ongoing refinement in training processes to better address the nuances of ethical reasoning and ensure more consistent alignment with human moral judgments. 

\subsection{Limitations}
In this investigation, we received our moral dilemmas from Scruples, who derived them from Reddit and annotated them using Mechanical Turk, which only encompasses specific types of moral ambiguity and isn't fully representative of real-world decision-making. This is attributable to the availability of existing datasets with human-annotated judgments in ambiguous situations. However, this scope permitted us to build a comprehensive understanding of LLM Calibration in a controlled setting with appropriate human annotations, which helped us gauge the models' performance. Future research could expand on the findings in this study by utilizing a dataset that captures diverse cultural and situational nuances, resulting in broader insights. 

The binary nature of human annotations also simplifies complex moral scenarios into 'right' or 'wrong'. While this expedited the process of measuring LLM Calibration, it reduces the degree of human reasoning. Although our approach provides a degree of alignment with human judgment, future studies should delve into more extensive evaluation methods that capture the full essence of moral ambiguity. 

\newpage
\bibliographystyle{plainnat}
\bibliography{refs}

\end{document}